\pdfoutput=1

\documentclass[11pt]{article}

\usepackage[]{ACL2023}
\usepackage{graphicx}
\usepackage{times}
\usepackage{multirow}
\usepackage{latexsym}
\usepackage{colortbl}
\usepackage{subfigure}
\usepackage{xcolor}
\usepackage{tabularx}
\usepackage{url}
\usepackage[T1]{fontenc}

\usepackage[utf8]{inputenc}

\usepackage{microtype}

\usepackage{inconsolata}

\title{CAISA at SemEval-2023 Task 8: Counterfactual Data Augmentation for Mitigating Class Imbalance in Causal Claim Identification}

\author{Akbar Karimi$^{1, 2}$ \and Lucie Flek$^{1, 2}$ \\
\small $^1$Conversational AI and Social Analytics (CAISA) Lab, University of Bonn, Germany \\
\small $^2$Conversational AI and Social Analytics (CAISA) Lab, University of Marburg, Germany \\
\small \texttt{\{akkarimi, lucie.flek\}@bit.uni-bonn.de} \\
      \small  \url{https://caisa-lab.github.io}}

\begin{document}
\maketitle
\begin{abstract}
The \texttt{class} \texttt{imbalance} problem can cause machine learning models to produce an undesirable performance on the minority class as well as the whole dataset. Using data augmentation techniques to increase the number of samples is one way to tackle this problem. We introduce a novel counterfactual data augmentation by verb replacement for the identification of medical claims. In addition, we investigate the impact of this method and compare it with 3 other data augmentation techniques, showing that the proposed method can result in a significant (relative) improvement in the minority class. 
\end{abstract}

\section{Introduction}

Automatic identification of medical claims \cite{khetan-EtAl:2023:SemEval, redhot23} is a task with various real-life applications in industries such as healthcare \cite{herland2017medical} and insurance \cite{wang2018leveraging} as well as content moderation \cite{schlicht2023multilingual}. 
However, it can be a difficult task due to the lack of data for all or some categories. 
One solution for such an issue is increasing the number of data points in each category, especially the one that has significantly fewer samples. We can do this using data augmentation techniques \cite{temraz2022solving}, which modify certain characteristics of an input sequence (or its representation in the embedding space) in order to create different versions of it. One example is entity replacement \cite{zeng-etal-2020-counterfactual}, where entities in one sequence can be swapped with equivalent ones from another sequence. The advantage of this type of augmentation is that it provides more real context to the target entities. 

Given that the task at hand is claim detection, we hypothesize that the verb in a sentence can be determinant in its category. Therefore, we address the problem of class imbalance using a novel data augmentation technique where we replace a verb in a sentence with other verbs from the training data. Our experiments show that verb replacement can improve the performance of a model on the target category. In addition, for more comparison, we experiment with several other data augmentation techniques, namely noise insertion \cite{karimi-etal-2021-aeda-easier}, entity replacement \cite{zeng-etal-2020-counterfactual}, augmentation with \texttt{YouChat}\footnote{\url{https://you.com/chat}}, and augmentation in the embedding space \cite{karimi2021adversarial}.
\section{Background} 
\textbf{Class Imbalance Problem.} This problem frequently comes up in many domains and applications. As a result, it has been tackled by a variety of methods such as oversampling \cite{ling1998data} and undersampling \cite{he2009learning}. The former method randomly selects some of the samples in the minority class and uses them multiple times for training in addition to the original samples. Contrarily, the latter randomly ignores some of the training examples from the majority class. However, the issue with them is that one (oversampling) might not always add new information to the training data, and the other (undersampling) might lose valuable information by not using some of the data points.

\noindent\textbf{Data Augmentation.} Another solution to tackling the class imbalance problem is to create synthetic instances from the existing ones \cite{chawla2002smote}. With this method, the resulting samples can be more diverse which can help avoid overfitting. However, the trade-off is that it can also introduce noise to the system although introducing noise is not always harmful \cite{karimi-etal-2021-aeda-easier}. In counterfactual data augmentation, words (or phrases) in a sentence are replaced with opposite (or different) ones from other sentences. This way, the focus parts of sentences are combined with different contexts, helping models in a better generalization to unseen sentences and combinations in the original data. For example, \citet{zeng-etal-2020-counterfactual} replace named entities from one sentence with another one to create new samples for the task of named entity recognition. We use the same approach for tackling \texttt{PIO} extraction (Subtask 2). To do so, we first create a dictionary of all the \texttt{PIO}s. Then, to augment a sentence in the training set, we replace its \texttt{PIO}s randomly with other ones from the dictionary. 

We compare the performance of the entity replacement with two other augmentation techniques. One is our counterfactual verb replacement method that we also used for Subtask 1 and the second one is an augmentation technique called \texttt{BAT} \cite{karimi2021adversarial} that takes place in the embedding space instead of the input space. 

\subsection{Data Exploration}
\begin{figure}
    \subfigure{\includegraphics[width=0.49\columnwidth]{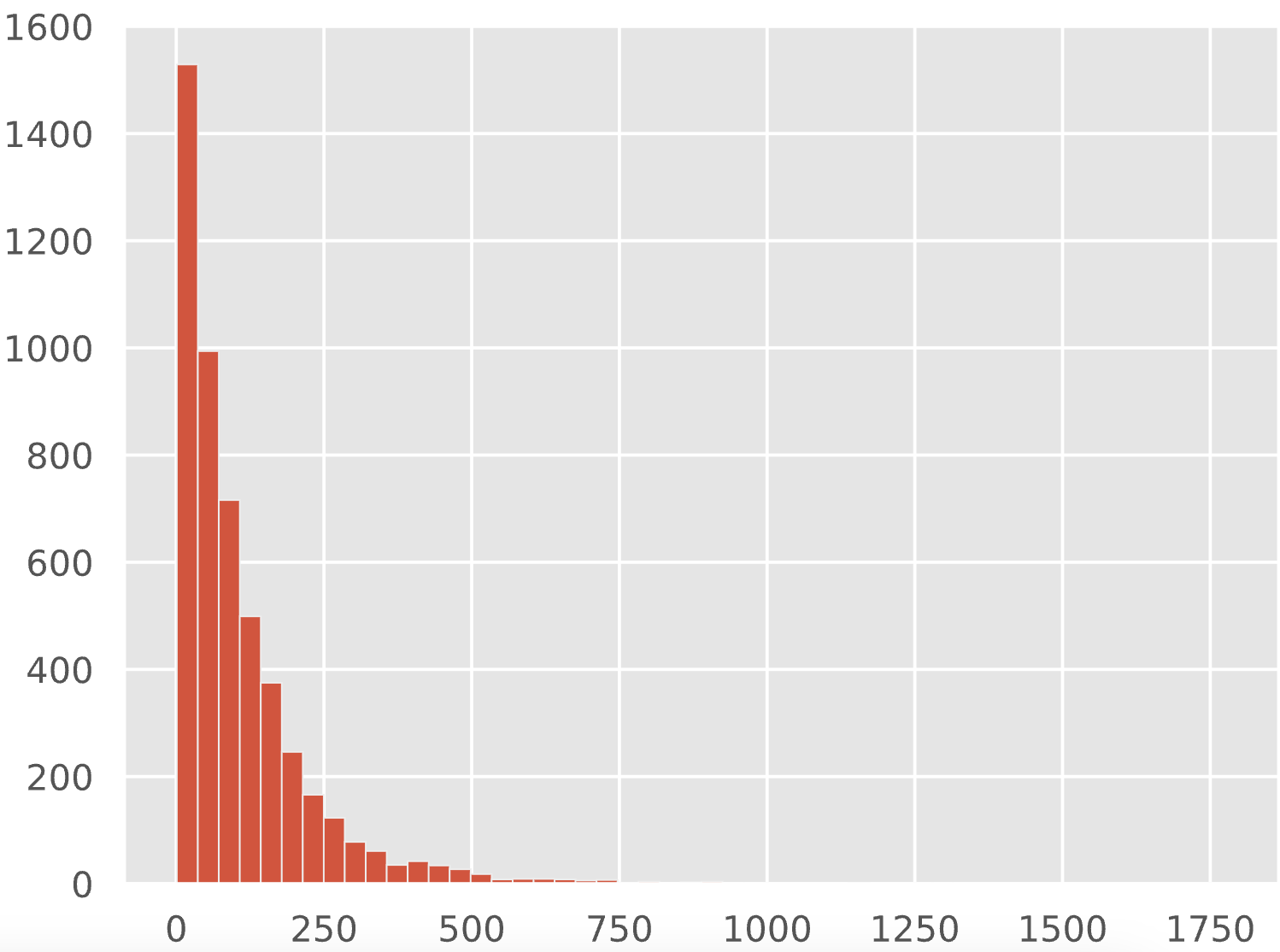}}
    \subfigure{\includegraphics[width=0.49\columnwidth]{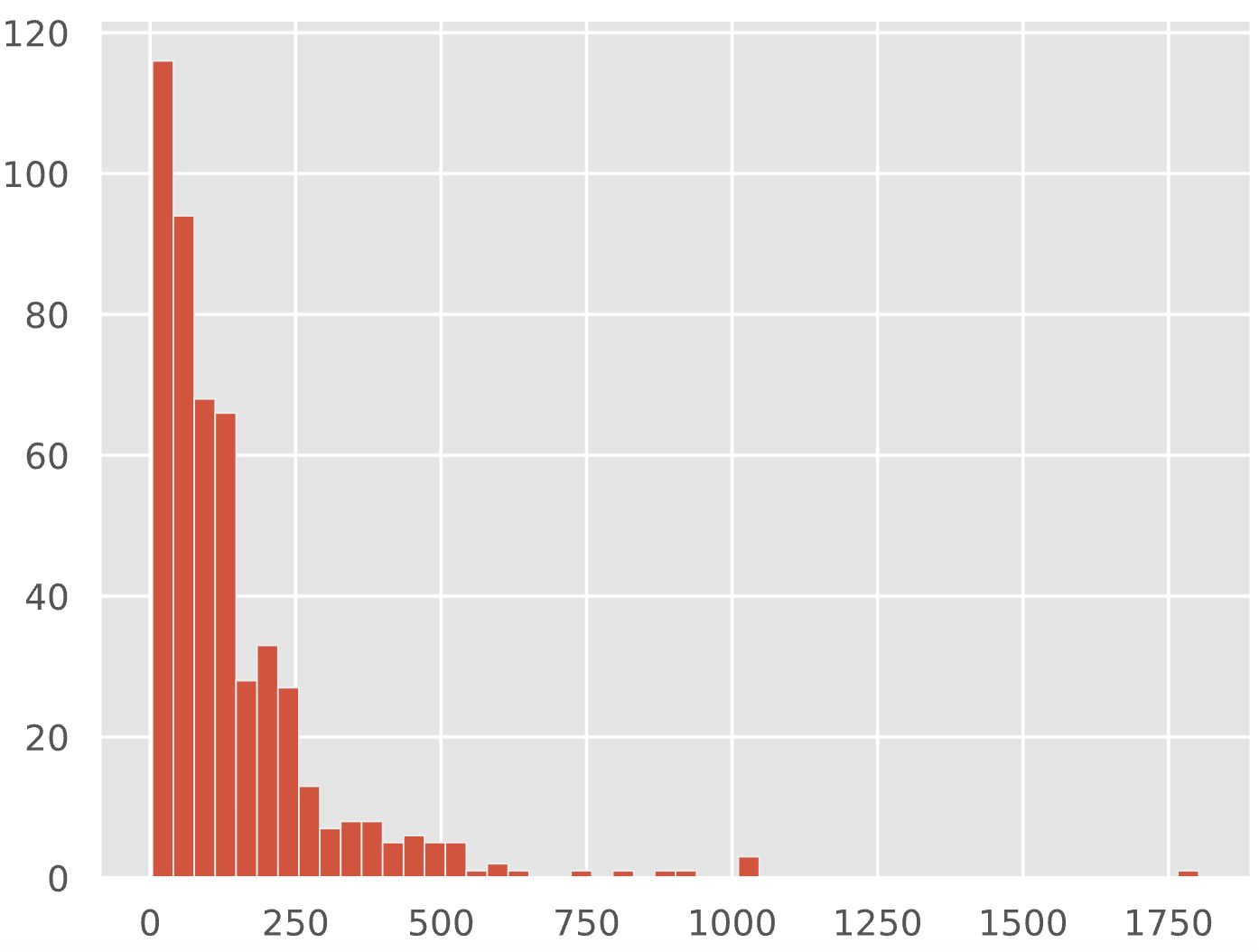}}
    \caption{Sentence length distributions in the training data for Subtasks 1 and 2.}
    \label{fig:sentence_length}
\end{figure}

The dataset \cite{redhot23} for the first task consists of \texttt{5710} texts that we split into two sets of training and development. Tables \ref{tab:sub1_statistics} and \ref{tab:sub2_statistics} show the number of samples for each set as well as the test set. As we can see from the tables, some texts can be longer than \texttt{1000} tokens. However, due to their low frequency (Figure \ref{fig:sentence_length}), we train our baseline (\texttt{DistilBERT}) with \texttt{512} tokens. 

\begin{table}
    \centering
    \begin{tabular}{l|c|c|c}
        Data & Texts & Unique words & Max length \\
        \hline
        Train & 5016 &39685&1777\\
        Dev & 694 &11287&1040\\
        Test & 1424 &16948&7876\\
    \end{tabular}
    \caption{Dataset statistics in Subtask 1}
    \label{tab:sub1_statistics}
\end{table}

\begin{table}
    \centering
    \begin{tabular}{l|c|c|c}
        Data & Texts & Unique words & Max length\\
        \hline
        Train & 501 &10834&1802\\
        Dev & 96  &3570&887\\
        Test &150 &4418&657\\
    \end{tabular}
    \caption{Dataset statistics in Subtask 2}
    \label{tab:sub2_statistics}
\end{table}

The categories in the dataset in Subtask 1 include \texttt{Claims (CLA)}, \texttt{Claim per experience (EXP)}, \texttt{Per experience (PER)}, and \texttt{Questions (QUE)}. Additionally, the dataset for Subtask 2 includes the categories of \texttt{Population (POP)}, \texttt{Intervention (INT)}, and \texttt{Outcome (OUT)}. Tables \ref{tab:sub1_detailed} and \ref{tab:sub2_detailed} show the number of tokens for each category in Subtasks 1 and 2, respectively. In the former, the claims (\texttt{CLA}) category has significantly fewer samples than other categories. In the latter, on the other hand, all the entities are vastly outnumbered by the outside (\texttt{O}) class. 

\begin{table}
    \centering
    \setlength{\tabcolsep}{4pt}
    \begin{tabular}{l|c|c|c|c|c}
        Data & \textbf{CLA} & \textbf{EXP} & \textbf{O} & \textbf{PER} & \textbf{QUE} \\
        \hline
        Train & 8183 & 33358 & 316676 & 138359 & 51707\\
        Dev & 1045 & 3995 & 44172 & 18298 & 6803 \\
        
    \end{tabular}
    \caption{Label distribution of train and dev data in Subtask 1.}
    \label{tab:sub1_detailed}
\end{table}

\begin{table}
    \centering
    \begin{tabular}{l|c|c|c|c}
        Data & \textbf{INT} & \textbf{O} & \textbf{OUT} & \textbf{POP} \\
        \hline
        Train & 800 & 69594 & 768 & 444 \\
        Dev & 185 & 12883 & 143 & 78 \\
    \end{tabular}
    \caption{Label distribution of train and dev data in Subtask 2}
    \label{tab:sub2_detailed}
\end{table}

\subsection{Data Augmentation Methods}
We experiment with four data augmentation methods for the identification of causal claims (Subtask 1) and three methods for \texttt{PIO} extraction (Subtask 2). Table \ref{tab:examples} shows an example for each method.

\begin{table*}
    \centering
    \begin{tabularx}{\textwidth}{|l|X|}
    \hline
        Original sentence & 80\% of people diagnosed with IBS have Sibo. \\
        \hline
        ER & 100 percent of people diagnosed with IBS have Sibo.\\
        VR (random) & 80 \% of people diagnosed with IBS cause Sibo .\\
        VR (antonym) & 80 \% of people diagnosed with IBS abstain Sibo .\\
        AEDA & 80\% of people diagnosed with IBS ! have Sibo.\\
        YouChat & Only a small fraction of those diagnosed with IBS actually have Small Intestinal Bacterial Overgrowth (SIBO).\\
        \hline
    \end{tabularx}
    \caption{An example augmentation from the \texttt{claims} class by the four augmentation methods, \texttt{ER} (entity replacement), \texttt{VR} (verb replacement), \texttt{AEDA} (an easier data augmentation).}
    \label{tab:examples}
\end{table*}

\noindent\textbf{AEDA} \cite{karimi-etal-2021-aeda-easier}. This approach is based on inserting punctuation marks into the input sentences. This can help improve the generalization capability of the model by changing the position of the words in the input sentence.

\noindent\textbf{Entity Replacement} \cite{zeng-etal-2020-counterfactual}. This method replaces the existing named entities with similar ones in the training dataset. In order to implement it, we first need to extract the entities from the sequences. We can do this using an off-the-shelf model for named entity recognition such as the \texttt{FLAIR} model \cite{akbik2018contextual}. Then, we create a dictionary of the named entities and in augmentation, we randomly choose one from the dictionary to replace with the original entity. One problem with this approach is that some sentences might not contain any entities. This will result in some sentences being repeated or discarded from the augmented batch.

\noindent\textbf{Verb Replacement}. Some verbs might be more indicative of the category to which they belong. To take advantage of this, we can create more diverse sequences by just replacing the verbs that are present in them. When replacing the verbs, we keep their tense intact. In addition, we experiment with two different ways for verb replacement. In one case, we first create a dictionary of verbs in the training data and select randomly from them when replacing a verb in a sentence. In the second case, we replace the verb with an antonym using WordNet \cite{miller1995wordnet}. The reason for not using the training data is that antonyms are rare and they might not be found in the
data.

\noindent\textbf{Augmentation with YouChat}. \texttt{YouChat} is a chatbot that can perform various guided actions such as augmenting sentences by producing contradictory ones. To do that we come up with a framing for our prompts that encourages diversity in the output as well as a contradiction. The reason for producing contradictory sentences is that, for the categories that, both the original and its contradiction can belong to the same category. For instance, for the \texttt{claims} category, if one sentence is considered to be a claim, then its contradiction can also be seen as a different claim. 

We use two prompts for pushing \texttt{YouChat} to produce diverse and counterfactual sentences: 1) \texttt{Contradict this sentence with colorful words "original sentence"}, and 2) \texttt{Without using despite, while, and although, contradict this sentence with colorful words "original sentence"}.
We use the first prompt to augment half of the sentences in the \texttt{claims} category. However, one problem that we notice with the outcome of this prompt is that after a couple of outputs, \texttt{YouChat} begins all sentences with expressions such as \texttt{although}, \texttt{despite}, and \texttt{while}. In order to change this, we augment the second half using the second prompt. This results in augmentations with different sentence structures.

\noindent\textbf{Augmentation with Adversarial Examples}. The BAT model \cite{karimi2021adversarial} trains the pre-trained language model in an adversarial manner where adversarial examples are created during the training in the embedding space. 

\begin{figure}
    \centering
    \includegraphics[scale=0.6]{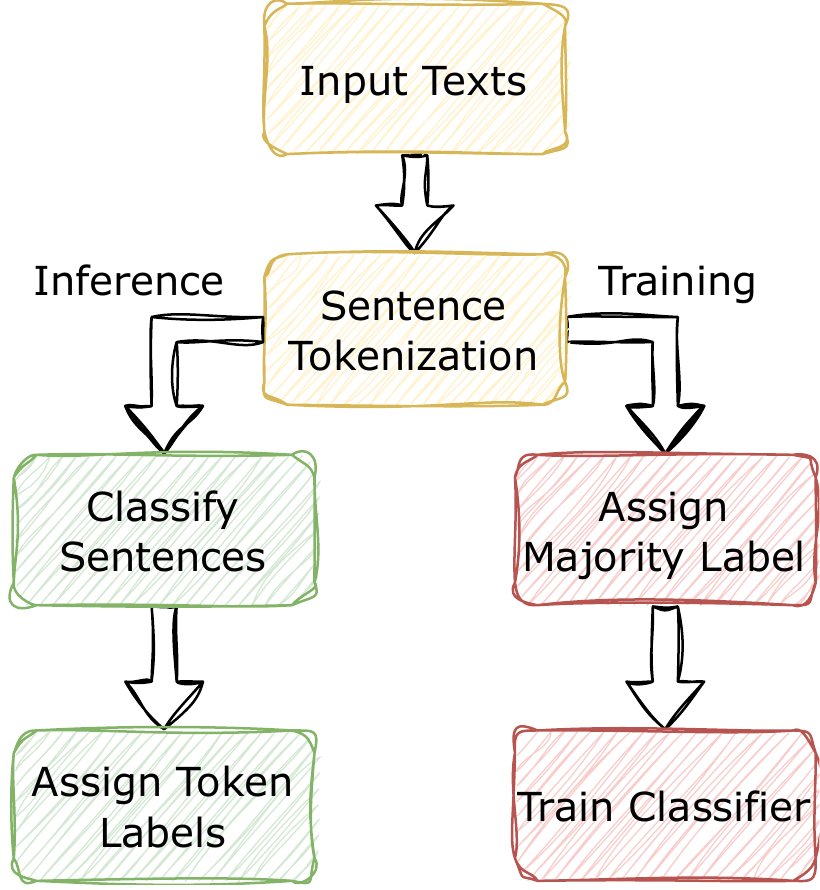}
    \caption{Workflow of our system\footnotemark}
    \label{fig:workflow}
\end{figure}
\footnotetext{Figure drawn using \url{https://draw.io}.}
\begin{table}
    \centering
    \begin{tabular}{l|c|c|c|c|c}
        Data & \textbf{CLA} & \textbf{EXP} & \textbf{O} & \textbf{PER} & \textbf{QUE}\\
        \hline
        Train & 401 & 1917 & 19826 & 7824 & 5064 \\
        Dev & 49 & 235 & 2666 & 995 & 633 \\
        
    \end{tabular}
    \caption{Number of sentences for each category after sentence tokenization in Subtask 1.}
    \label{tab:sentences}
\end{table}

\begin{table*}
\setlength{\tabcolsep}{4pt}    
\centering
    \begin{tabular}{l|l|c|c|c|c|c|c|c|c}
         &\textbf{Method} & \textbf{CLA} & \textbf{EXP} & \textbf{O} & \textbf{PER} & \textbf{QUE} & \textbf{Precision}& \textbf{Recall}& \textbf{F1}\\
         \hline
         \multirow{2}*{Baseline} &CRF & 11.1 & 12.7 & 76.5 & 49.5 & 71.0 & 51.1 & 42.1 & 44.1 \\
         &DistilBERT & 25.8 & 32.0 & 77.6 & 56.8 & 76.4 & 57.9 & 52.0 & 53.7 \\
         \hline
         
         \multirow{5}*{400}&CRF + ER & 13.7 & 12.1 & 77.4 &48.2 & 70.3 & 56.1 & 41.7 & 44.3 \\
         &CRF + VR (random) & 11.3 & 15.1 & 76.4 & 50.2 & 69.5 & 50.5 & 42.4 & 44.5 \\
         & CRF + VR (antonym) & 11.7 & 8.9 & 75.9 & 49.6 & 69.2 & 47.2 & 41.6 & 43.1\\
         &CRF + AEDA & 5.4 & 4.8 & 77.0 & 47.8 & 67.7 & 44.5 & 39.5 & 40.6 \\
         &CRF + YouChat &9.2 & 12.0 & 75.3 & 49.1 & 68.7 & 46.1 & 41.4 & 42.9\\
         \hline
         \multirow{5}*{100}&DB + ER & \colorbox{lime}{27.9} & 29.3 & 77.5 & 56.5 & 76.8 & 56.7 & 52.3 & 53.6 \\
         &DB + VR (random) & \colorbox{lime}{27.9} & \colorbox{cyan}{34.2} & 77.5 & 56.2 & 76.6 & 58.3 & 52.8 & \colorbox{lime}{54.5} \\
         &DB + VR (antonym) & 27.5 & 29.1 & 77.6 & 56.3 & 77.2 & 57.7 & 52.0 & 53.5\\
         &DB + AEDA & 25.5 & 28.5 & 77.8 & 56.7 & 76.9 & 56.3 & 51.9 & 53.1 \\
         &DB + YouChat & 26.2 & \colorbox{lime}{35.8} & 77.6 & 56.9 & 76.1& 59.0 & 52.9 & \colorbox{lime}{54.5} \\
         \hline
         \multirow{5}*{400}&DB + ER & \colorbox{lime}{29.7} & 29.8 & 77.1 & 57.1 & 77.2 & 56.7 & 53.1 & \colorbox{cyan}{54.2} \\
         &DB + VR (random) & 22.3 & 29.9 & 77.7 & 56.5 & 77.2 & 56.5 & 51.3 & 52.7\\
         &DB + VR (antonym) & \colorbox{pink}{18.6} & 32.1 & 77.4 & 56.6& 76.3 & 57.0 & 50.7 & \colorbox{pink}{52.2}\\
         &DB + AEDA & \colorbox{cyan}{28.9} & \colorbox{lime}{35.2} & 77.6 & 56.7 & 76.7 & 59.1 & 53.1 & \colorbox{lime}{55.0} \\
         &DB + YouChat & 21.0 & 30.1 & 77.0 & 57.0 & 76.3 & 54.8 & 51.1 & 52.3\\
    \end{tabular}
    \caption{Subtask 1. Experiments with \texttt{100} and \texttt{400} augmented samples for the \texttt{claims} class with \texttt{DistilBERT} \texttt{(DB)} and \texttt{CRF} models using Verb Replacement \texttt{(VR)}, Entity Replacement \texttt{(ER)}, \texttt{AEDA}, and \texttt{YouChat} augmentations. Green shows the best performer, blue is the second best, and red is the worst.}
    \label{tab:st1_all}
\end{table*}

\section{System Overview}
The annotated data gives us the span of each category. The spans can be complete sentences or part of a sentence. One approach to address the task is to frame it as a token classification task, similar to named entity recognition tasks. However, named entities seem to be easier to spot because of their locality. On the contrary, given that a longer sequence of words could belong to a category, we take a broader look to recognize them. As a result, we formulate the problem as sentence classification. The dataset statistics indicate that in more than \texttt{87} percent of the resulting sentences, all the words have the same label. 

\subsection{Workflow}
As can be seen in Figure \ref{fig:workflow}, we first split the texts into sentences using a sentence tokenizer toolkit from the \texttt{NLTK} library \cite{loper2002nltk}. Then, for training, we assign the label of the majority of the tokens to the sentence and train the model with the resulting data. For the inference part, after sentence tokenization, we classify each sentence using the trained model and assign the sentence label to the individual tokens.

\subsection{Sentence-Tokenized Dataset Statistics}

Separating the input texts into sentences results in just over \texttt{35K} sentences which are distributed heavily in favor of the outside (\texttt{O}) class. Table \ref{tab:sentences} shows the statistics of the resulting data. As we can see, only a small proportion of the sentences belong to the claims category. We augmented the samples in this category to mitigate the class imbalance. 

\section{Baseline Models}

We compare the performance of augmentation methods with two baseline models, namely a conditional random fields \cite{lafferty2001conditional} model and the \texttt{DistilBERT} pre-trained language model for Subtask 1 \cite{sanh2019distilbert}, and the \texttt{BioBERT} model \cite{lee2020biobert} for Subtask 2. 

\noindent\textbf{Conditional Random Fields (CRF)}. This model is particularly suited for sequence labeling tasks. It considers a set of manually defined feature functions to predict the label of a token. In our case, we only consider some simple features such as the word itself, the word endings, whether the word is uppercase or lowercase, and if it is a number or not. With the same features, we also consider bigrams. 

\noindent\textbf{DistilBERT}. This model is a lighter and more robust version of the \texttt{BERT} model \cite{devlin2019bert}. We also consider the performance of this model without any augmentation as one of the baselines.

\noindent\textbf{BioBERT}. This is another variant of the \texttt{BERT} model that has been trained on medical texts in addition to the general text used for training \texttt{BERT}. 

\section{Results and Analysis}
We perform augmentation on the \texttt{claims} class for Subtask 1 and the whole dataset for Subtask 2. 

\subsection{Subtask 1: Causal Claim Identification}
\begin{table*}
\centering
    \begin{tabular}{l|c|c|c|c|c|c|c|c}
         \textbf{Method} & \textbf{CLA} & \textbf{EXP} & \textbf{O} & \textbf{PER} & \textbf{QUE} & \textbf{Precision}& \textbf{Recall}& \textbf{F1}\\
         \hline
         DistilBERT & 25.8 & 32.0 & 77.6 & 56.8 & 76.4 & 57.9 & 52.0 & 53.7 \\
         \hline
         DB + ER & 23.8 & 27.8 & 76.6 & 55.1 & 77.1 & 54.7 & 51.1 & 52.1\\
         DB + VR (random) & \colorbox{pink}{15.4} & \colorbox{lime}{36.5} & 77.7 & 56.3 & 77.4 & 57.7 & 51.2 & 52.7 \\
         DB + AEDA & 24.4 & 32.1 & 76.9 & 55.6 & 76.9 & 57.7 & 51.6& 53.2\\
    \end{tabular}
    \caption{Subtask 1. Results with 4 augmentations for all \texttt{400} samples in \texttt{claims} class with \texttt{DistilBERT} \texttt{(DB)} using Verb Replacement \texttt{(VR)}, Entity Replacement \texttt{(ER)}, and \texttt{AEDA} methods.}
    \label{tab:st1_multiple}
\end{table*}

For this task, the \texttt{CRF} model provides a relatively well-performing baseline despite its simplicity. Notably, from Table \ref{tab:st1_all}, we can see that it does well on the \texttt{QUE} class with \texttt{71} percent. This can be attributed to the fact that the \texttt{CLA} class is easier to detect although the number of samples in this category is a lot lower than the \texttt{PER} class that has a performance of \texttt{49.5} percent. Quite understandably, the lowest performing classes were the \texttt{CLA} and \texttt{EXP} classes, which can be due to having only a small number of samples in addition to their difficulty. 

\texttt{DistilBERT}, on the other hand, shows an almost \texttt{10} percent overall improvement as well as on individual classes over the \texttt{CRF} model, which is expected given the large number of parameters it has compared to \texttt{CRF} ($\approx$\texttt{68M} vs. $\approx$\texttt{10M}).

\noindent\textbf{Impact of augmentation on CRF}. The impact of augmentations with the \texttt{CRF} model is somewhat mixed. While \texttt{ER} helps the \texttt{claims} class improve by two percent, others show no improvement or negative improvement. It is possible that the \texttt{CRF} model is more vulnerable to out-of-distribution changes.

\noindent\textbf{Impact of augmentation on DistilBERT}. Considering the effect of the augmentation methods on the overall performance of \texttt{DistilBERT}, we experiment with two scenarios: first with \texttt{100} augmented sentences and then with \texttt{400} augmentations. Table \ref{tab:st1_all} shows that, in the first case, \texttt{VR} \texttt{(random)} and \texttt{YouChat} have helped the model improve by almost one point while the \texttt{ER} method has had a slightly negative effect. The effect on the minority class, however, was positive for all the augmentation methods except for \texttt{AEDA}, with \texttt{ER} and \texttt{VR} \texttt{(random)} showing more than two percent improvement and \texttt{YouChat} \texttt{0.4} percent. With \texttt{400} augmentations, we see that only \texttt{ER} and \texttt{AEDA} have improved the class performance. This can be attributed to the increase in the amount of noise as we include more augmentations.

\noindent\textbf{Impact of multiple augmentations}. In this experiment, we investigate how multiple augmentations can impact the \texttt{DistilBERT} model on the studied dataset. Therefore, for each tokenized sentence, we produce four augmentations. We do this only for \texttt{ER}, \texttt{VR} \texttt{(random)}, and \texttt{AEDA} since for \texttt{YouChat} the manual work is time-consuming and for \texttt{VR} \texttt{(antonym)}, there is only one antonym for a verb. As we can see from Table \ref{tab:st1_multiple}, more augmentations of the \texttt{claims} class have a negative effect on the class itself while improving the results on the \texttt{EXP} (claim per experience) class. Given that this class also contains claims, it seems that more data for the \texttt{claims} class could also help \texttt{claims} \texttt{per} \texttt{experience} class. 

\subsection{Subtask 2: PIO Frame Extraction}

\begin{table}
    \centering
    \begin{tabular}{l|c|c|c}
        \textbf{Method} &  \textbf{Precision} & \textbf{Recall} & \textbf{F1} \\
        \hline
        BioBERT & 47.2 & 11.7 & 18.8\\
        \hline
        BioBERT + ER &  32.5 & 11.7 & 17.2 \\
        BioBERT + BAT & 20.8 & 17.7 & 19.1 \\
        BioBERT + VR & 25.7 & 16.4 & 20.1 \\
    \end{tabular}
    \caption{Subtask 2. Results with one augmentation using \texttt{ER}, \texttt{BAT}, and \texttt{VR} with \texttt{BioBERT}. We augment the entire dataset.}
    \label{tab:st2_all}
\end{table}

For this experiment, we utilized the \texttt{BioBERT} model \cite{lee2020biobert} as the baseline. Table \ref{tab:st2_all} shows the effect of three augmentation methods on this model with \texttt{100} examples augmented from the \texttt{claims} category. As we can see, overall, verb replacement is more effective than other methods although entity replacement makes more sense for this task since in \texttt{ER}, we increase the number of sentences using similar entities. This should provide a more diverse context for the existing entities in the training data. 

\section{Conclusion}
We proposed verb replacement as a novel counterfactual data augmentation technique to increase the number of samples in the minority class for causal claim identification. Then, we showed that this method can significantly improve the performance of the machine learning model in the minority class. Comparing it with three other augmentation methods, we also found out that the proposed method can outperform them in some cases.

\bibliography{anthology,custom}
\bibliographystyle{acl_natbib}




\end{document}